\documentclass[sigconf]{acmart}

\usepackage{graphicx}
\usepackage{amsmath, xspace,  comment}
\usepackage{tikz, xcolor}
\usepackage{algpseudocode}
\usepackage[linesnumbered]{algorithm2e}
\usepackage{colortbl}
\usepackage{textcase} 
\usepackage{enumitem}
\usepackage{balance}
\usepackage{graphicx}
\usepackage{multirow}
\usepackage{hhline}
\usepackage{caption}
\usepackage{cite}
\usepackage[caption=false]{subfig}
\usepackage{soul}
\usepackage{hyperref}
\usepackage{makecell}

\AtBeginDocument{%
  }

\newcommand*\cib[1]{\tikz[baseline=(char.base)]{
                            \node[shape=circle,fill=black,text=white,draw,inner sep=0.3pt] (char) {#1};}}

\newcommand{\ie}{{\em i.e.,}\xspace}

\newcommand{\BfPara}[1]{{\noindent\underline{\bf#1.}}\xspace}

\usepackage{tcolorbox}
\tcbuselibrary{theorems}

\newtcbtheorem[]{observation}{\textbf{Observation}}%
{colback=white,colframe=white!10!black,fonttitle=\bfseries,coltitle=white}{}

\copyrightyear{2021} 

\hypersetup{
	colorlinks=true,
	urlcolor=blue!70!black,
	linkcolor=red!70!black,
	citecolor=purple,
} 

\begin{document}
\fancyhead{}
\title{Unveiling Vulnerabilities in Interpretable Deep Learning Systems with Query-Efficient Black-box Attacks}

%
\author{Eldor Abdukhamidov}
\email{abdukhamidov@skku.edu}
\affiliation{%
  \institution{Computer Science and Engineering\\Sungkyunkwan University}
  \city{Suwon-si}
  \state{Gyeonggi-do}
  \country{South Korea}
}

\author{Mohammed Abuhamad}
\email{mabuhamad@luc.edu}
\affiliation{%
  \institution{Department of Computer Science\\Loyola University Chicago}
  \city{Chicago}
  \state{Illinois}
  \country{USA}
}

\author{Simon S. Woo}
\email{swoo@g.skku.edu}
\affiliation{%
  \institution{Department of Artificial Intelligence\\Sungkyunkwan University}
  \city{Suwon-si}
  \state{Gyeonggi-do}
  \country{South Korea}
}

\author{Eric Chan-Tin}
\email{chantin@cs.luc.edu}
\affiliation{%
  \institution{Department of Computer Science\\Loyola University Chicago}
  \city{Chicago}
  \state{Illinois}
  \country{USA}
}

\author{Tamer Abuhmed}
\email{tamer@skku.edu}
\affiliation{%
  \institution{{Computer Science and Engineering}\\Sungkyunkwan University}
  \city{Suwon-si}
  \state{Gyeonggi-do}
  \country{South Korea}
}


\begin{abstract}
Deep learning has been rapidly employed in many applications revolutionizing many industries, 
but it is known to be vulnerable to adversarial attacks. Such attacks pose a serious threat to deep learning-based systems compromising their integrity, reliability, and trust. Interpretable Deep Learning Systems (IDLSes) are designed to make the system more transparent and explainable, but they are also shown to be susceptible to attacks.
In this work, we propose a novel microbial genetic algorithm-based black-box attack against IDLSes that requires no prior knowledge of the target model and its interpretation model. The proposed attack is a query-efficient approach that combines transfer-based and score-based methods, making it a powerful tool to unveil IDLS vulnerabilities. Our experiments of the attack show high attack success rates using adversarial examples with attribution maps that are highly similar to those of benign samples which makes it difficult to detect even by human analysts. Our results highlight the need for improved IDLS security to ensure their practical reliability.
\end{abstract}




\maketitle

\section{Introduction}

The rapid development and deployment of deep learning models have led adversaries to exploit vulnerabilities in the application pipeline to compromise results or lead models to misbehave \citep{qirui2022mlxpack, sun2022leveraging,zhang2020interpretable}. Studies have shown that deep neural network models are susceptible to adversarial examples, which are carefully designed samples used for adversarial purposes such as poisoning, evasion, model extraction, and inference \citep{abdukhamidov2023microbial,abdukhamidov2023single}.

Interpretable Deep Learning Systems (IDLSes) are deep learning models with interpretable knowledge representations. They have been shown to be more robust against adversarial attacks, as interpretation can reveal adversarial manipulations, \ie the added perturbations to the example input. However, recent studies have shown that IDLSes in white-box settings are still susceptible to adversarial manipulations \citep{zhang2020interpretable,10.1145/3488932.3527283,abdukhamidov2021advedge,abdukhamidov2022interpretations}. To be specific, adversarial samples can mislead the target deep learning model and deceive its coupled interpreter simultaneously.

Although attacks in white-box scenarios are based on complete knowledge of the target model and can achieve a high attack success rate with high confidence, they are impractical in most circumstances. In contrast, black-box attacks assume that the adversary can only query the model and access the output without extended knowledge of the system's components or the model's parameters, and are therefore more realistic. Transfer-based and score-based attacks are common examples of this type of attack \citep{juraev2022depth,harvey2009microbial, alzantot2019genattack, abdukhamidov2023single}.

Attacking IDLSes in black-box settings is still an unexplored field with many challenges. This work proposes a black-box attack that generates adversarial examples to mislead the target models and their coupled interpreters. The proposed approach is gradient-free and query-efficient based on transfer-based and score-based attacks. We evaluated our approach against two deep learning models and one interpreter on the ImageNet dataset and show the possibility and practicality of generating malicious examples with arbitrary predictions and carefully manipulated interpretations in order to achieve a high attack success rate in a black-box environment.

\vspace{6ex}
\BfPara{Contributions} Our contributions can be summarized as follows:
\begin{itemize} [leftmargin=2em]
\item We propose the black-box version of AdvEdge attack \citep{abdukhamidov2021advedge} to generate adversarial samples against IDLSes.
\item We empirically evaluate the effectiveness of the attack from the perspective of two deep learning models and one interpretation model. Based on experimental results, we show that the proposed approach achieves a high attack success rate with a smaller number of queries to attack several target deep learning models and their interpreters on the ImageNet dataset.
\end{itemize}

\vspace{1ex}
\BfPara{Organization} The remainder of the paper is organized as follows: Section~\ref{sec:methods} describes the notations and terms used in the paper and presents the proposed attack and its underlying mechanisms; Section~\ref{sec:evaluation} provides the results of empirical evaluations of attack effectiveness and robustness against deep learning and interpretation models; Section~\ref{sec:related} surveys recent research studies in the domain; Section~\ref{sec:conc} concludes the paper.

\section{Methods} \label{sec:methods}
The section describes the proposed attack in black-box settings with a detailed explanation of the methods adopted. 

\subsection{Concepts and Notations} \label{sec:fundamental}
The notation, terms, and symbols used in the paper are introduced in this subsection.  

\vspace{1ex}
\BfPara{Classifier} This work focuses on image classification using two types of deep neural network models: white-box and black-box. In a black-box setting, we denote the target model as $f(x)=y \in Y$, where $y$ is a single category from a set of categories $Y$. In a white-box setting, we denote the source model as $f'(x)=y \in Y$.

\vspace{1ex}
\BfPara{Interpreter} We use an existing interpretation model $g$ to generate an interpretation map $m$ that displays the feature importance for a sample $x$ classified by $f$: $g(x; f)= m$. Our approach uses post-hoc interpretability \citep{dabkowski2017real, fong2017interpretable, karpathy2015visualizing, murdoch2018beyond}, which requires another model to interpret the decision process of the current classification model.

\vspace{1ex}
\BfPara{Adversarial Attack} PGD attack generates an adversarial sample $\hat{x}$ to make the source model $f'$ misclassify $\hat{x}$ into another category: $f'(\hat{x}) \neq y$. It works by perturbing the input pixels and is implemented using a projection operator $\prod$, a learning rate $\alpha$, a loss function $\ell_{adv}$, and a norm ball $\mathcal{B}_{\varepsilon}(x)$ with a range $\varepsilon$. The update rule is as follows.

\begin{equation*}
\hat{x}^{(i+1)} = \prod {\mathcal{B}{\varepsilon}(x)}\left(\hat{x}^{(i)} - \alpha . ~sign(\nabla_{\hat{x}}\ell_{adv}(f'(\hat{x}^{(i)})))\right)
\end{equation*}

\vspace{1ex}
\BfPara{Threat Model} We consider a black-box setting in which the adversary has limited access to the target deep learning classifier ($f$) but no access to the interpretation model ($g$), which is a realistic scenario for the attack.


\subsection{Attack Formulation} \label{sec:attack_formulation}

To effectively attack IDLSes, it is necessary to deceive both the deep learning model and its interpretation model. AdvEdge \citep{abdukhamidov2021advedge} presents a technique for generating an adversarial sample $\hat{x}$ that satisfies four critical conditions. These conditions include: \cib{1} successfully tricking the deep learning classifier $f'$, \cib{2} producing an interpretation map $\hat{m}$ similar to the benign sample $x$, \cib{3} being visually imperceptible, and \cib{4} limiting noise to the edge of the sample.
The attack framework can be summarized as follows.

\begin{equation*}
\begin{split}
    \min _{\hat{x}} : \Delta(\hat{x}, x) \quad s.t. \left\{ \begin{array}{lcr}
         f'(\hat{x}) \neq y, \quad s.t. \quad \| \hat{x} - x\|_{\infty} \in \{-\epsilon, \epsilon\}\\
         g(\hat{x}; f') = \hat{m}, \quad s.t. \quad \hat{m} \cong m \\
         \Delta(\hat{x}, x) \sim edge(x \cap m)
         \end{array}\right.
\end{split}
\end{equation*}

By using this formulation, the adversarial sample generated ensures that the predicted category is different from the original one, the interpretation map remains similar to the benign sample, and the added perturbation is limited to the sample's edges that intersect with the interpretation map.
To achieve this, the attack framework minimizes the overall adversarial loss ($\ell_{adv}$) that includes both the classification loss $\ell_{prd}(f'(x)) = - log(f'(x))$ and the interpretation loss $\ell_{int}(g(x;f', m) = |g(x;f') - m|_{2}^2$.

The overall adversarial loss is formulated as follows:

\begin{equation*} \label{eq:mainFormula}
    \ell_{adv}= \min_{\hat{x}} \ell_{prd}(f'(\hat{x})) + \lambda ~\ell_{int}(g(\hat{x};f'), m)
\end{equation*}
where the hyper-parameter $\lambda$ balances $\ell_{prd}$ and $\ell_{int}$.

The final adversarial framework can be described as follows: 

\begin{equation*} 
\begin{split}
    \hat{x}^{(i+1)} = \prod _{\mathcal{B}_{\varepsilon}(x)}\left(\hat{x}^{(i)} - N_{w}~ \alpha .~ sign(\nabla_{\hat{x}}\ell_{adv}(\hat{x}^{(i)}))\right)
\end{split}
\end{equation*}

In the above equation, $\prod$ represents the production operator, $\mathcal{B}_{\varepsilon}$ is a norm ball, $\alpha$ is the learning rate, $x$ denotes the input sample, and $\hat{x}^{(i)}$ denotes the adversarial sample generated at the \textit{i}-th iteration.
Furthermore, the edge operator function $N_{w}$ is used to optimize the location and magnitude of the added perturbation:

\begin{equation*}
\begin{split}
d = \sqrt{d_{h}^2 + d_{v}^2}\\
    N_{w} = d \cap m
\end{split}
\end{equation*}
where $d$ is an image that contains edges of the sample $x$ extracted through the formula $d = \sqrt{d_{h}^2 + d_{v}^2}$, where $d_{h}$ and $d_{v}$ represent the horizontal and vertical edge information of the sample. The attack process utilizes the intersection of the edges of a sample image and its interpretation map to identify critical regions.

The PGD framework \citep{madry2017towards} is employed to generate initial adversarial samples for the genetic algorithm in a white-box setting, using a transfer-based approach to attack the source deep learning models and their interpreters. Additionally, the Microbial Genetic Algorithm (MGA) \citep{abdukhamidov2023microbial,harvey2009microbial} is utilized to optimize adversarial samples against the black-box deep learning classifier $f'$.

\subsection{MGA}
MGA \citep{harvey2009microbial} is a genetic algorithm that leverages a gradient-free optimization technique to generate candidate solutions. The algorithm operates by iteratively evolving a set of samples, referred to as a population, to produce optimal candidates with higher fitness scores. Each iteration, or generation, involves the evaluation of the quality of each member of the population through a fitness function that assigns a value based on a defined objective function of the optimization process.

The fitness function plays a crucial role in determining the likelihood of a particular sample being selected for the next generation through a process that involves crossover and mutation. Samples that demonstrate high fitness scores are more likely to be selected for this process, and the iterative evolution of the population continues until an optimal candidate that satisfies the problem's objective function is found.

MGA is a useful optimization technique in scenarios where the objective function is unknown or difficult to compute, as it enables the exploration of the search space without relying on gradients. The approach is particularly effective in solving complex optimization problems with a large number of variables. More details can be found in Section \ref{subsec: blackbox_attack}.

\subsection{Black-box Implementation} \label{subsec: blackbox_attack}

Our approach is based on transfer-based learning techniques \citep{szegedy2013intriguing, dong2019evading}. We generate adversarial samples against a deep learning model in a white-box setting and use them as the initial population for MGA. MGA updates the initial population to produce new generations of adversarial examples to deceive a black-box deep learning model $f'$ and mislead its coupled interpreter $g$.

The attack consists of genetic algorithm operators: initialization, selection, crossover, mutation, and population update. Seeding the initial population with an optimal solution helps the technique converge fast. We evaluate each individual in the population by applying a loss function as the fitness function. Unlike the traditional genetic algorithms, the selection process of our method randomly picks two samples, one with a higher fitness score and the other with a lower fitness score, to keep the perturbation in the newly generated sample area that is considered important by the target model and its interpreter. We generate new offspring by transferring the genetic data of the winner and the loser with the predefined crossover rate. Mutation diversifies the population and introduces enough diversity to reach points outside the regions of the local optima.

Overall, the AdvEdge algorithm is effective in generating adversarial samples against a source deep learning model and its interpreter. 
The generated samples are then used as seeds for the initial population. The fitness scores of the population are evaluated by sending them to the target model in a black-box setting. If the attack requirements are met, the algorithm stops further steps. Otherwise, the algorithm repeats the steps until it succeeds or reaches the query threshold.

\section{Experiments and Evaluation} \label{sec:evaluation}
In this section, we provide detailed information on the settings and metrics used for our experiment to measure the performance of the proposed attack in terms of the deep learning and interpretation models given in the paper.

\subsection{Experimental Settings}
\vspace{1ex}
\BfPara{Datasets} We conducted our experiment on the ImageNet dataset \citep{deng2009imagenet}, consisting of 1.2 million images for 1,000 categories. To evaluate our attack, we randomly select one image from each category in the ImageNet validation set \citep{deng2009imagenet}, resulting in a total of 1,000 images. We ensure that the selected images are correctly classified by the target model $f$ with a classification confidence score greater than 60\%. For this experiment, we set the value of $\epsilon$ at 8, which is similar to the settings used in the AdvEdge attack \citep{abdukhamidov2021advedge}, and represents the perturbation scale in the range of [0, 225]. Through these experiments, we validate that our proposed attack is effective across all categories of the selected deep learning models.

\vspace{1ex}
\BfPara{Classifiers} The experiment conducted in this study involves the use of two popular models, namely DenseNet-169 and ResNet-50, which were pre-trained on the ImageNet dataset. These models were used as both the source and target models for the transfer-based attack, which involved generating adversarial samples that could be transferred from the source model to the target model. The attack was limited to a maximum of 50,000 queries and the step size $\alpha$ and the number of iterations were set at 1/255 and 300, respectively. These values were selected based on the settings used in a previous attack called AdvEdge \citep{abdukhamidov2021advedge}. By using these pre-trained models and established attack settings, we aim to evaluate the effectiveness of our proposed attack in a controlled and reproducible manner.

\vspace{1ex}
\BfPara{Interpreters} CAM \citep{zhou2016learning} interpreter is adopted as the representative of the interpretation models. CAM utilizes the feature maps of the convolutional layers in a deep learning model to generate interpretation maps: $m_{c} = \sum_{i} w_{i, c} a_{i} (j, k)$, where $a_{i}(j, k)$ is the activation of the \textit{i}th channel at the spatial location $(j, k)$ and $ w_{i, c}$ is the weight of the \textit{i}-th input and the \textit{c}-th output in the linear layer of a deep learning model. We set $\lambda$ in Equation ($\ell_{adv}= \min_{\hat{x}} \ell_{prd}(f'(\hat{x})) + \lambda ~\ell_{int}(g(\hat{x};f'), m)$) at 0.204 for the CAM that is found effective in AdvEdge \citep{abdukhamidov2021advedge}. We use its open-source implementations for the experiment.

\subsection{Attack Evaluation}

We evaluate the proposed attack using various metrics to answer the following questions: \cib{1} \textit{Is it effective against black-box deep learning models?} \cib{2} \textit{Can it deceive interpretation models by generating interpretation maps similar to benign samples?} \cib{3} \textit{Is it effective against defensive black-box deep learning models?} \cib{4} \textit{Can it deceive interpretation models with defensive black-box deep learning models?}

\vspace{1ex}
\BfPara{Evaluation Metrics} Different evaluation metrics are used to assess the effectiveness of the proposed attack against both deep learning classifiers and interpreters.

For deep learning classifiers, the following metrics are used:

\begin{itemize}[leftmargin=2em]
\item \textbf{Attack success rate}: It calculates the ratio of successful attack cases to total attack cases.
\item \textbf{Average queries}: The metric evaluates the efficiency of the attack algorithm in generating successful adversarial examples in a black-box setting.
\item \textbf{Noise rate}: The metric is used to evaluate the quality of the adversarial examples generated by the attack algorithm.
\end{itemize}

For interpreters, the following metrics are used:

\begin{itemize}[leftmargin=2em]
\item \textbf{Qualitative comparison}: The metric evaluates the similarity between the interpretations of adversarial images and their benign counterparts.
\item \textbf{IoU Test}: The metric measures the similarity of interpretation maps using the Intersection-over-Union score for different threshold values.
\end{itemize}

\begin{figure*}[h]
    \centering
    \captionsetup{justification=justified}
    \includegraphics[width=0.95\linewidth]{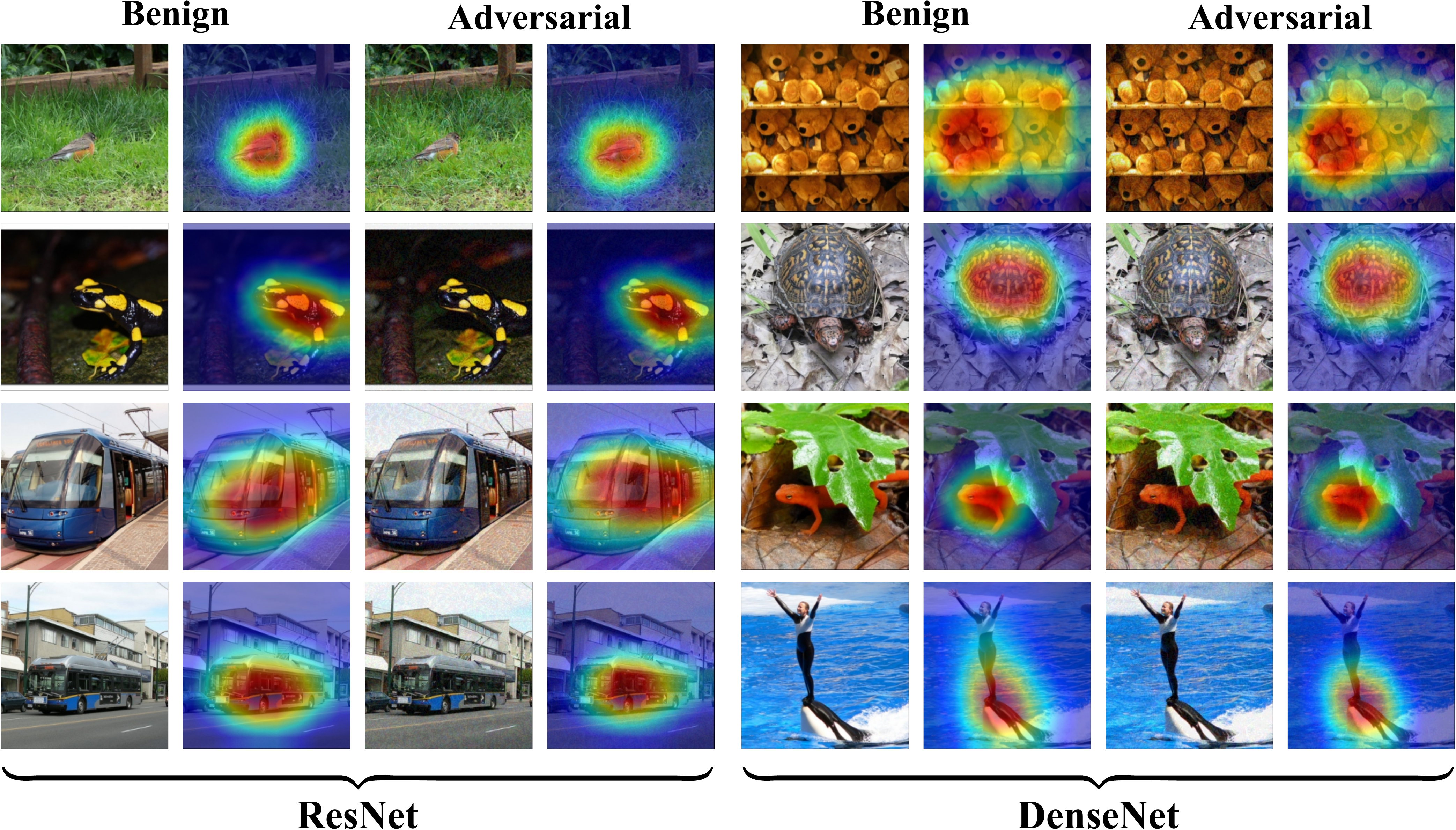}
    \caption{Attribution maps of benign and adversarial samples generated by our attack using CAM on ResNet and DenseNet.}
    \label{fig:blackbox_examples}
\end{figure*}

\begin{table}
\centering
\caption{Success rate, average queries, and average noise of the proposed attack against different classifiers and interpreters using 1,000 images. The attack is based on black-box settings.}
\label{tab:ASR}
\resizebox{\linewidth}{!}{%
\begin{tabular}{|c|c|c|c|c|c|} 
\toprule
\rowcolor[rgb]{1,1,1} \begin{tabular}[c]{@{}>{\cellcolor[rgb]{1,1,1}}c@{}}\\ \textbf{Interpreter}\end{tabular} & \makecell{\textbf{Source} \\ \textbf{Model}} & \makecell{\textbf{Target}\\ \textbf{Model}} & \makecell{\textbf{Success}\\ \textbf{Rate}} & \makecell{\textbf{Average} \\ \textbf{Queries}} & \makecell{\textbf{Average}\\ \textbf{Noise Rate}}  \\ 
\midrule
\multirow{2}{*}{\textbf{CAM}}                                                                                                          & \textbf{ResNet}       & DenseNet              & 0.99                  & 209.76                & 0.20 $\pm$ 0.06           \\ 
\cline{2-6}
    & \textbf{DenseNet}     & ResNet                & 1.00                  & 188.53                & 0.20 $\pm$ 0.06           \\
\bottomrule
\end{tabular}
}
\end{table}

\subsection{The Attack against Deep Learning Models}
\label{sec:attack_dnns}
In this section, we present the evaluation of the effectiveness of our proposed adversarial attack on two popular model architectures, namely ResNet and DenseNet. To assess the efficacy of our attack, we implemented and tested it on two interpretable machine learning models using the Class Activation Mapping (CAM) technique with ResNet and DenseNet as source models.

The results of our experiments on the aforementioned scenarios are reported in Table \ref{tab:ASR}. Our attack on DenseNet models achieved an impressive attack success rate of 0.99 and was found to be highly query efficient, requiring an average of only 209.76 queries. The average noise rate in the target model was stable at $0.20$ $\pm$ $0.06$.
For the CAM interpreter with ResNet, our attack achieved a complete attack success rate on the target model with an average of 188.53 queries. The average noise rate remained stable. These results show improved performance with more complex source models.

Our results indicate that the proposed attack is highly effective against popular deep learning architectures and interpretable machine learning models, highlighting potential security vulnerabilities and the need for robust defenses against adversarial attacks.      

\begin{figure}[t]
    \centering
    \captionsetup{justification=justified}
    \includegraphics[width=0.95\linewidth]{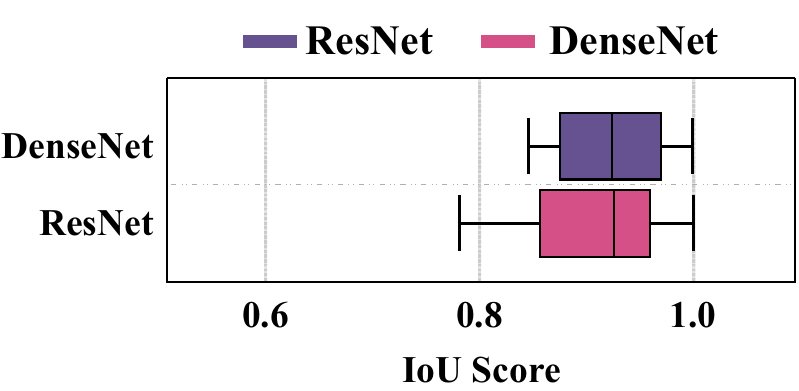}
    \caption{IoU scores of interpretation maps generated by our attack using CAM interpreter and ResNet, DenseNet as source models. y-axis represents the target models}
    \label{fig:blackbox_advedge_iou}
\end{figure}

\begin{observation}{The Attack against Deep Learning}{}
  Our proposed attack has demonstrated a high degree of effectiveness in deceiving deep learning models, achieving a consistently high attack success rate across different source models. This highlights the significant potential for our attack to pose a threat to the security and reliability of deep learning models.   
\end{observation}

\subsection{The Attack against Interpreters} \label{sec:attack_int}
In this section, we investigate the effectiveness of our proposed attack on the similarity between benign and adversarial interpretations, using a qualitative comparison and the Intersection over Union (IoU) test.

\vspace{1ex}
\BfPara{Qualitative comparison} Our qualitative comparison of the attribution maps generated by the Class Activation Mapping (CAM) interpreter for benign and adversarial samples showed that it was difficult to differentiate between the two. Our findings indicate that the adversarial attribution maps generated by our attack are highly reliable and comparable in quality to those produced from benign inputs. Figure \ref{fig:blackbox_examples} shows several examples of comparison between benign and adversarial samples generated by the attack. 

\vspace{1ex}
\BfPara{IoU Test} To further assess the similarity between the two types of attribution maps, we used the IoU test, which measures the overlap between two maps. Our attack achieved highly balanced IoU scores on different deep learning models with the CAM interpreter, indicating that the adversarial attribution maps generated by our attack are similar to the benign attribution maps. Figure \ref{fig:blackbox_advedge_iou} summarizes our attack performance, which can be considered significant for a black-box attack.

Our findings highlight the effectiveness of our proposed attack in generating adversarial interpretation maps that are similar to their benign counterparts, raising concerns about the security and reliability of interpretable machine learning models.

\begin{observation}{The Attack against Interpreters}{}
  Our proposed attack has been shown to generate adversarial interpretation maps that are visually similar to their corresponding benign counterparts. This characteristic makes it challenging to distinguish between adversarial and benign maps, highlighting the potential for our attack to undermine the reliability and trustworthiness of interpretable machine learning models.   
\end{observation}

\section{Related Work} \label{sec:related}
This section comprehensively reviews prior research on attacks targeting deep neural network models. The survey encompasses studies conducted on white-box and black-box attacks, utilizing diverse techniques such as transfer-based attacks, interpretation-based attacks, and gradient-based attacks.

\vspace{1ex}
\BfPara{Transfer-based attacks} In the realm of attacks against deep learning models, transfer-based attacks have received considerable attention in previous research. These attacks utilize adversarial samples generated by white-box attacks against one model to attack other black-box models. The potential effectiveness of transfer-based attacks has been demonstrated in various studies \citep{szegedy2013intriguing, huang2019enhancing, dong2019evading, abdukhamidov2022black}. For instance, researchers have proposed methods to enhance the transferability of adversarial samples by adding perturbations to the hidden layers of a model or convolving the gradient via a specific kernel. These studies highlight the importance of considering transfer-based attacks when assessing the robustness of models and offer insight into techniques to improve the transferability of adversarial samples.

\vspace{1ex}
\BfPara{Interpretation-based attacks} This part discusses interpretation-based adversarial attacks that can deceive both the target deep learning models and their interpreters \citep{zhang2020interpretable}. A recent study proposed white-box attacks called AdvEdge and AdvEdge$^+$ against deep learning models and their interpreters, highlighting the vulnerability of models that rely on interpretable features for decision making \citep{abdukhamidov2021advedge}. These attacks show the importance of considering the interpretability of deep learning models in addition to their accuracy and robustness. Furthermore, the proposed attacks provide a valuable tool for evaluating the interpretability of models and assessing their susceptibility to adversarial attacks.

\vspace{1ex}
\BfPara{Gradient-free attacks} Heuristic methods, including evolution strategies and genetic algorithms, have been utilized to create adversarial attacks that can generate visually imperceptible samples to deceive deep learning models \citep{alzantot2019genattack}. GenAttack is a gradient-free optimization attack that can generate adversarial samples against black-box models with fewer queries. Another study proposed a query-efficient attack called MGAAttack \citep{wang2020mgaattack}, which uses transfer-based techniques to improve its efficacy. These attacks show the susceptibility of deep learning models to adversarial attacks and highlight the need to develop more robust defense mechanisms to enhance their security. By analyzing these attacks, researchers can identify weaknesses in models and devise better defenses against adversarial attacks.

\section{Conclusion} \label{sec:conc}
In this study, we propose a black-box version of the AdvEdge attack that can effectively deceive deep learning models and their interpreters. Our attack combines transfer-based and score-based methods to generate adversarial examples that are difficult for the target models to classify correctly while also producing adversarial interpretation maps that are highly similar to the corresponding benign interpretations. Furthermore, our attack is both gradient-free and query-efficient, making it suitable for practical scenarios where access to model parameters or gradients may be limited. We evaluated the effectiveness of our proposed attack on various deep learning models, including ResNet and DenseNet, and their interpreters, such as the Class Activation Mapping (CAM) interpreter. Our experimental results show that our attack achieves a high success rate in deceiving target models and interpreters. Moreover, we performed a qualitative comparison and an Intersection over Union (IoU) test to evaluate the similarity between adversarial and benign interpretation maps. Our comparison of the attribution maps generated by the CAM interpreter for both benign and adversarial samples showed that it was difficult to distinguish between them. These results suggest that our attack can generate adversarial interpretation maps with a level of reliability that is comparable to that of benign inputs. In general, our proposed attack highlights the importance of developing more robust deep learning models and interpretability techniques to enhance their security against adversarial attacks. Furthermore, our work underscores the need to develop effective defense mechanisms that can detect and prevent such attacks in real-world scenarios.
\balance
\bibliographystyle{ACM-Reference-Format}
\bibliography{ref}

\end{document}